\documentclass[lettersize,journal,twocolumn]{IEEEtran}
\usepackage[T1]{fontenc}
\usepackage[utf8]{inputenc}
\usepackage{pdflscape}
\usepackage{multirow}
\usepackage{array}
\usepackage{comment}
\usepackage{makecell}
\usepackage{tabu}
\usepackage{tabularx}

\usepackage[numbers]{natbib} 
\usepackage{booktabs} 
\usepackage{amsmath}
\usepackage{hyperref}
\usepackage{mathtools}
\usepackage[capitalize]{cleveref}

\usepackage{listings}
\usepackage{tikz}
\usetikzlibrary{matrix,backgrounds,shapes.multipart,positioning,calc,chains,arrows,arrows.meta,decorations.pathmorphing,tikzmark,shadows,trees,mindmap,backgrounds}
 
\usepackage{pgfplots}
\pgfplotsset{%
    ,compat=1.12
    ,every axis x label/.style={at={(current axis.right of origin)},anchor=north west}
    ,every axis y label/.style={at={(current axis.above origin)},anchor=north east}
    }

\def\BibTeX{{\rm B\kern-.05em{\sc i\kern-.025em b}\kern-.08em
    T\kern-.1667em\lower.7ex\hbox{E}\kern-.125emX}}
\usepackage{balance}

\makeatletter
\pgfmathdeclarefunction{erf}{1}{%
  \begingroup
    \pgfmathparse{#1 > 0 ? 1 : -1}%
    \edef\sign{\pgfmathresult}%
    \pgfmathparse{abs(#1)}%
    \edef\x{\pgfmathresult}%
    \pgfmathparse{1/(1+0.3275911*\x)}%
    \edef\t{\pgfmathresult}%
    \pgfmathparse{%
      1 - (((((1.061405429*\t -1.453152027)*\t) + 1.421413741)*\t 
      -0.284496736)*\t + 0.254829592)*\t*exp(-(\x*\x))}%
    \edef\y{\pgfmathresult}%
    \pgfmathparse{(\sign)*\y}%
    \pgfmath@smuggleone\pgfmathresult%
  \endgroup
}
\makeatother

\begin{document}
\title{Interpretable Symbolic Regression for Data Science: Analysis of the 2022 Competition}

\author{
    \IEEEauthorblockN{
        F. O. de Franca\IEEEauthorrefmark{1},
        M. Virgolin\IEEEauthorrefmark{2},
        M. Kommenda\IEEEauthorrefmark{5}, 
        M. S. Majumder\IEEEauthorrefmark{3}, 
        M. Cranmer\IEEEauthorrefmark{6},
        G. Espada\IEEEauthorrefmark{8}, 
        L. Ingelse\IEEEauthorrefmark{8}, 
      }
      \\
    \IEEEauthorblockN{
        A. Fonseca\IEEEauthorrefmark{8}, 
        M. Landajuela\IEEEauthorrefmark{9}, 
        B. Petersen\IEEEauthorrefmark{9},
        R. Glatt\IEEEauthorrefmark{9}, 
        N. Mundhenk\IEEEauthorrefmark{9}, 
        C. S. Lee\IEEEauthorrefmark{9},
        J. D. Hochhalter\IEEEauthorrefmark{10}, 
    }
    \\
    \IEEEauthorblockN{
        D. L. Randall\IEEEauthorrefmark{10},
        P. Kamienny\IEEEauthorrefmark{11},
        H. Zhang\IEEEauthorrefmark{12},
        G. Dick\IEEEauthorrefmark{13}, 
        A. Simon\IEEEauthorrefmark{14}, 
        B. Burlacu\IEEEauthorrefmark{5},
        Jaan Kasak\IEEEauthorrefmark{15}, 
    }
    \IEEEauthorblockN{
        Meera Machado\IEEEauthorrefmark{15},
        Casper Wilstrup\IEEEauthorrefmark{15},
        W. G. {La Cava}\IEEEauthorrefmark{3}
    }
    \thanks{
    \IEEEauthorrefmark{1}Center for Mathematics, Computation and Cognition (CMCC), Heuristics, Analysis and Learning Laboratory (HAL), Federal University of ABC, Santo Andre, Brazil. 
    }
    \thanks{\IEEEauthorrefmark{2}Evolutionary Intelligence group, Centrum Wiskunde \& Informatica, Science Park 123, Amsterdam, Netherlands.}
    \thanks{\IEEEauthorrefmark{3}Computational Health Informatics Program, Boston Children's Hospital, Harvard Medical School, Boston, USA. 
    }
    \thanks{\IEEEauthorrefmark{5}Heuristic and Evolutionary Algorithms Laboratory (HEAL), University of Applied Sciences Upper Austria, Hagenberg, Austria.}
    \thanks{\IEEEauthorrefmark{6}Center for Computational Astrophysics, Flatiron Institute and Department of Astrophysical Sciences of Princeton University, USA.}
    \thanks{\IEEEauthorrefmark{8}LASIGE, Faculdade de Ciências, Universidade de Lisboa, Lisboa, Portugal. 
    }
    \thanks{\IEEEauthorrefmark{9}Computational Engineering Division, Lawrence Livermore National Laboratory, Livermore, USA. 
    }
    \thanks{\IEEEauthorrefmark{10}University of Utah, Department of Mechanical Engineering, Utah, USA.}
    \thanks{\IEEEauthorrefmark{11}Meta, FAIR, France.}
    \thanks{\IEEEauthorrefmark{12}Victoria University of Wellington, School of Engineering and Computer Science, New Zealand.}
    \thanks{\IEEEauthorrefmark{13}University of otago, Department of Information Science, New Zealand.}
    \thanks{\IEEEauthorrefmark{14}
        Institut für Angewandte Physik,
        Universität Tübingen;
        Max Planck Institute for Intelligent Systems,
        Tübingen, Germany
    }
    \thanks{\IEEEauthorrefmark{15}
        Abzu AI,
        Orient Plads 1, Nordhavn 2150, Denmark
    }
    \thanks{Preprint Under Review.  
    Corresponding author: W. G. {La Cava} (email: william.lacava@childrens.harvard.edu)}
}




\maketitle

\begin{abstract}
    Symbolic regression searches for analytic expressions that accurately describe studied phenomena. 
    The main attraction of this approach is that it returns an interpretable model that can be insightful to users.
    Historically, the majority of algorithms for symbolic regression have been based on evolutionary algorithms. 
    However, there has been a recent surge of new proposals that instead utilize approaches such as enumeration algorithms, mixed linear integer programming, neural networks, and Bayesian optimization.
    In order to assess how well these new approaches behave on a set of common challenges often faced in real-world data, we hosted a competition at the 2022 Genetic and Evolutionary Computation Conference consisting of different synthetic and real-world datasets which were blind to entrants.  
    For the real-world track, we assessed interpretability in a realistic way by using a domain expert to judge the trustworthiness of candidate models. 
    We present an in-depth analysis of the results obtained in this competition, discuss current challenges of symbolic regression algorithms and highlight possible improvements for future competitions.
\end{abstract}



\IEEEpeerreviewmaketitle

\section{Introduction}~\label{sec:introduction}

\IEEEPARstart{T}he goal of \textbf{symbolic regression} (SR)~\cite{koza1992genetic} is to find a parameterized function $f(x, \theta)$ in the form of an analytic expression that best fits the given data. 
Such expressions may describe non-linear interactions with a similar accuracy to that of opaque models (e.g., deep neural networks or tree ensembles), while making it possible for humans to understand their behavior in great detail like transparent models do (e.g., linear models or decision trees).
Consequently, SR has been used to uncover new phenomena from collected data, extending our knowledge of physics~\cite{cranmer2020discovering}, chemistry~\cite{hernandez2019fast}, and other  fields.

Despite its utility, the SR problem is NP-hard~\cite{virgolin2022symbolic}, consequently, a number of different algorithms have been proposed to tackle SR from different communities, from evolutionary computation, to operations research, to deep learning~\cite{petersen2019deep,biggio2021neural,kamienny2022end}.
In an effort to bridge existing gaps between the communities and draw a picture of the state of the art in SR, \emph{SRBench}\footnote{\url{https://cavalab.org/srbench}} was recently proposed~\cite{srbench}.
SRBench is an extensive benchmark that includes different SR methods as well as other machine learning methods, and evaluates expression accuracy, simplicity, and the algorithm's capability of re-discovering the data-generating function on physical problems. 
The results put forward by SRBench showed that SR algorithms based on genetic programming (GP) tend to be the best performing at the moment on a variety of real-world tabular problems. 
Moreover, it was shown that SR is itself a viable approach to modelling tabular data, since in many cases the discovered expressions exhibited similar test accuracy to state-of-the-art ML methods, commonly accepted to be the best for this task~\cite{SHWARTZZIV202284}. 

Notwithstanding the importance of such results, a number of interesting questions on SR remain unanswered to date.
For example, SRBench did not include a study on how well each SR algorithm behaved on some common challenges of real-world data, such as presence of noise, redundant features, or extrapolation behaviour outside of the training distribution.
To shed light on these open questions, we organized a competition named \emph{Interpretable Symbolic Regression for Data Science} hosted at the 2022 Genetic and Evolutionary Computation Conference (GECCO). 
The goal of this competition was to promote the study of multiple imporant facets of SR.
To that end, we considered a number of assessment criteria that go beyond expression accuracy and length: 
\emph{feature selection}, \emph{sensitiveness to local optima}, \emph{accuracy in the extrapolation regime}, \emph{sensitivity to noise}, \emph{rediscovery of known laws} and, importantly, \emph{interpretability} according to an expert of the field for a real-world use, i.e., forecast of key indicators for the COVID-19 pandemic.
The competition received $11$ distinct submissions, based on GP, deep learning, and combinations thereof.

In this paper, we describe the rationale behind our design choices for the competition, perform an in-depth analysis, and draw findings based on the obtained results.
Also, we report the current challenges for the SR community and give some insights for future improvements of this competition. The paper is organized as follows. 
In Section~\ref{sec:sr} we give a brief introduction to SR and the current state-of-the-art. Section~\ref{sec:srbench} we describe the competition rules, tasks, evaluation and competitors.  In Section~\ref{sec:results} we show and analyse the obtained results in the competition with focus on the winning entries. Finally, Section~\ref{sec:discussion} discuss the results summarizing the insights obtained from the competition and we emphasize the current challenges of SR research. We conclude this paper with some final thoughts and future steps in Section~\ref{sec:conclusion}.

\section{Background}\label{sec:sr}


SR is  the problem of searching for a closed-form expression
, data by composing simpler functions (often called primitives) from an user-defined set. This search is usually guided by a loss function $\mathcal{L}(y,f(x, \theta))$ that measures the approximation error of the regression model and the measured target.


The main advantages of creating an analytic expression are the possibilities for manual inspection, debugging, and adaptation by the practitioner. 
This manual manipulation can lead to a better understanding of the model, improve the accuracy of the model, or allow specialists to incorporate expert knowledge. 
Apart from the manual manipulation, we can also analyse the behavior of the model using common regression tools~\cite{friedman2001greedy,bomarito2022automated}. 
We remark that simpler models, such as analytic expressions, can be competitive with opaque models in terms of accuracy on many datasets 
~\cite{srbench,rudin2019stop}. Specifically for SR, there is evidence that analytic expressions excel at extrapolating outside the training data region in comparison to gradient boosting and neural networks~\cite{RolandExtrapolate}.
Although SR has some advantages over other regression methods, until recently, the adoption of SR in practice has been limited for a number of reasons:

\begin{itemize}
    \item longer training time 
    compared to traditional regression; 
    \item a lack of implementations in easy-to-use toolboxes
    \item the difficulty of customizing existing toolboxes; 
    \item a lack of supporting tools to interpret generated models;
    \item a smaller and more insular research community compared to other machine learning fields; and
    \item a lack of shared dependencies / platforms between methods and toolboxes. 
\end{itemize}

Attempts to address these limitations began with~\cite{orzechowskiWhereAreWe2018b} and led to the creation of SRBench~\citep{srbench}.
SRBench not only revealed the benefits of using SR for prediction tasks but also created an environment that enables practitioners easy access to different SR approaches using a \emph{scikit-learn}-like Python interface. 



\section{The 2022 Symbolic Regression Competition}\label{sec:srbench}

The SRBench competition of 2022 was held at the GECCO 2022 conference in Boston, MA. The goal of this competition was to assess different aspects of SR algorithms other than prediction accuracy, to understand the current challenges of SR, and also to further stimulate the growth of SRBench, which is intended to be a reference and ever-improving benchmark for tracking the state-of-the-art in the field.

The structure of the competition included a qualifying stage to filter methods that did not meet a minimal level of performance, defined as the accuracy of a simple linear regressor. 
After the qualifying stage, the competition had two tracks: a synthetic track to assess different properties of interest measured against ground-truth solutions, and a real-world track, where an expert was asked to judge competing models of COVID-19 spread using publicly available health data. Qualifying methods were evaluated on both tracks, and separate prizes were designated for each.

For each run, a candidate SR algorithm had a pre-specified time budget of $1$ hour for datasets up to $1000$ samples and $10$ hours for datasets up to $10000$ samples. Candidates were  responsible for ensuring that the runtime of their algorithm (including choices of hyperparameter optimization) would not exceed the budget.
In order to make comparisons as fair as possible, participants were not given advance notice of what datasets would be used for the competition. 
They knew only that datasets would follow the format in the Penn Machine Learning Benchmark (PMLB)~\cite{romanoPMLBV1Opensource,olsonPMLBLargeBenchmark2017}, and that datasets from PMLB would be used in the qualifying stage. 

With the objective of making the competition accessible and reproducible by external peers, we created a new branch in the SRBench repository\footnote{\url{https://github.com/cavalab/srbench/tree/Competition2022}} with instructions on how to add a new method and how to run the competition on a local machine.

\subsection{Qualification Track}

In the qualification track, we used a selection of $20$ PMLB datasets to verify whether the competitors were capable of finding better models than a linear model baseline. 
The main objective of this track was to filter entrants that did not adhere to a minimum acceptable prediction accuracy. 
The evaluation criteria for this track was the median $R^2$ score on $10$ distinct runs. 
We ranked the results by the median across the different datasets and disqualified those approaches that were ranked lower than using plain linear regression.
The chosen datasets are available at the supplementary material.

\subsection{Synthetic tracks}

For the synthetic tracks, our goal was to evaluate different tasks that simulate challenges often observed in real-world data. By evaluating such challenges, we hope to gain a more granular understanding of specific tasks for which current algorithms excel or have room for improvement.

In these tracks, we used multiple evaluation criteria: $R^2$, simplicity, and a task-specific score for each tasks. We then computed the aggregated rank as the harmonic means of the ranks for each criterion for $n$ different data-sets.
The harmonic mean imposes that, to be highly ranked, you cannot have a low rank in any of the criteria. 
This avoids the situation that an SR algorithm returns a very simple model with low $R^2$ and still ranks high among the competitors.

The simplicity score is defined as $\operatorname{round}(-\log_5(s), 1)$, where $s$ is the number of nodes in the expression tree after being simplified by \emph{sympy}~\cite{sympy}, and $\operatorname{round}(x, n)$ rounds the value $x$ to the $n$-th place.
Rounding was introduced to provide some tolerance for similarly-sized expressions. 


We tested $5$ different tasks in this track of the competition:

\begin{enumerate}
    \item \textbf{Rediscovery of the exact expression:} the SR model must match the exact generating function of the dataset.
    \item \textbf{Selection of relevant features:} the SR model must only use relevant features, discarding any noisy feature.
    \item \textbf{Escaping local optima:} the SR model should be composed of noise-free low-level features  instead of noisy interactions of the original features. 
    \item \textbf{Extrapolation accuracy:} given a dataset with a limited range on the variables' domains, the SR model should behave as intended outside this range.
    \item \textbf{Sensitivity to noise:} given different levels of noise applied to the target value, the SR algorithm should recover an expression close to the original function.
\end{enumerate}

In the following subsections we will explain each one of these tasks in further details with the data generating process and evaluation criteria.

\subsubsection{Rediscovery of exact expression}

In this task, the SR algorithm should return a model that is within a constant (multiplicative or additive) factor of the generating function. 
To assess this, we rely on the ability of \emph{sympy} to manipulate expressions algebraically and evaluate the equivalence of two expressions. 

\begin{table}[t!]
    \centering
    \caption{Generating functions for the rediscovery of exact expression task and their corresponding degree of difficulty.}
    {
    \tabulinesep=1.1mm
    \begin{tabu}{c|c|c}
    \hline 
    Function & Difficulty     &  Generating Function \\
    \hline 
    $f_1(x)$ & Easier   &  $0.4 x_1 x_2 - 1.5 x_1 + 2.5 x_2 + 1$ \\
    $f_2(x)$ & Easy     &  $f_1(x) + \log{(30 x_3^2)}$ \\
    $f_3(x)$ & Medium & $\frac{f_1(x)}{0.2(x_1^2 + x_2^2) + 1}$ \\
    $f_4(x)$ & Hard & $\frac{f_1(x) + 5.5 \sin{(x_1 + x_2)}}{0.2(x_1^2 + x_2^2) + 1}$ \\ 
    \hline 
    \end{tabu}
    }
    \label{tab:exact-eq}
\end{table}

In this task we created three levels of difficulty that are created from the same base function by including increasingly complex terms. 
Table~\ref{tab:exact-eq} shows the corresponding generating functions.

\subsubsection{Selection of relevant features}

For this task, we created datasets with $20$ features following the generating function:
\begin{equation}
\small
    0.11 x_1^3 + 0.91 x_3 x_5 + 0.68 x_7 x_9 + 0.26 x_{11}^2 x_{13} + 0.16 x_{15} x_{17} x_{19}.
    \label{eq:select-feat}
\end{equation}

As we can see from Eq.~\ref{eq:select-feat}, the variables with an even index are not used when generating the dataset (they are random variables unrelated to the function). The main goal is that the SR model identifies and use only the odd indexed features. For this purpose, we introduce the specific score for this task given by:

\begin{equation}
    \operatorname{feature-select}(f_{true}, f_{pred}) = \frac{\operatorname{true-feats} - \operatorname{false-sel-feats}}{\operatorname{false-feats}},
    \label{eq:feature-select}
\end{equation}
where $\operatorname{false-feats}$ is the set of random valued features and $\operatorname{false-sel-feats}$ is the set of those noisy features used by the SR model.
There are also three difficulty levels in this task -- easy, medium, hard -- that are created by adding different level of noise to the target value and replacing the target with:

\begin{equation}
    y_{noise} = \mathcal{N}\left(y, \sigma_y \sqrt{\frac{\operatorname{ratio}}{1 - \operatorname{ratio}}}\right)
    \label{eq:noise}
\end{equation}
where $\operatorname{ratio}$ is the noise level. The ratio of each difficulty setting is $0.025$ for easy, $0.05$ for medium, and $0.1$ for hard.

\subsubsection{Escaping local optima}

For this task, the dataset contains the original features and some meta-features that are higher-level building blocks of the generating function.
However, crucially, the data regarding the meta-features contains noise.
Therefore, a good but suboptimal model can be constructed by combining the meta-features instead of the original features.
We defined the meta-features (without noise) as:

\begin{align}
    f(x) = \underbrace{0.77 x_1 x_2}_{g_1(x)} + \underbrace{1.52 x_2 x_3}_{g_2(x)} + \underbrace{1.2 x_4^2}_{g_3(x)} + \underbrace{0.31 x_1 x_4 x_5}_{g_4(x)} 
    \label{eq:local-optima} \\
    + \underbrace{0.23 x_3 x_4 x_5}_{g_5(x)}. \nonumber
\end{align}
and the generating function with $n$ meta-features is given by $f_n(x) = \sum_{i=1}^{n}{g_i(x)}$.

The corresponding dataset is built as a list of samples in the format $[x_1, x_2, x_3, x_4, x_5, g_1 + \epsilon, \ldots, g_n + \epsilon, f_n(x)]$, where $\epsilon$ is an additive noise following Eq.~\ref{eq:noise} with a ratio of $0.1$. 
Since noise is added to the data of the meta-features $g$, the optimal fit to $f_n$ can only be recovered if the meta-features are ignored, and the original features $x$ are used.
There are also $3$ levels of difficulty controlled by the value of $n$: easy with $n=3$, medium with $n=4$, hard with $n=5$. The task-specific score used here is the same as the one for the feature selection task.

\subsubsection{Extrapolation accuracy}

The goal of this tasks is to evaluate the performance of SR models when the validation set is outside the training data domain. This is a tricky situation as by fitting the model on a limited space there can be many equally good models that behaves differently outside the training boundaries.
Models are thus rewarded for choosing a minimally complex hypothesis among those describing the training data.  
For this task we used the generating function:

\begin{equation}
    f(x) = \operatorname{erf}(0.22 x) + 0.17 \sin{(5.5 x)},
\end{equation}
where $\operatorname{erf}$ is the error function defined as

\begin{equation*}
    \operatorname{erf}(z) = \frac{2}{\sqrt{\pi}}\int_{0}^{z}{e^{-t^2}dt}.
\end{equation*}

Besides not having an analytic solution to this problem, the training and validation datasets were split such as $-15 \leq x \leq 15$ for the training set and $15 < x \leq 40$ for the validation set. 

The specific measure for this task is the evaluation of the $R^2$ score on the validation data. Like in the other tasks, we created three different level of difficulties by adding noise following Eq.~\ref{eq:noise} with ratios $0.05$ (easy), $0.1$ (medium), and $0.2$ (hard).

\subsubsection{Sensitivity to Noise}

In this task the SR models are created using a noisy training data and evaluated on a noiseless validation data. The generating function used was:

\begin{equation*}
    f(x) = \frac{0.11 x^4 - 1.4 x^3}{0.68 x^2 + 1}.
\end{equation*}

The different noise ratios were $0.05$ (easy), $0.1$ (medium), and $0.15$ (hard). 

\subsection{Real-world track}

\begin{figure}
    \centering
    \includegraphics[width=\linewidth]{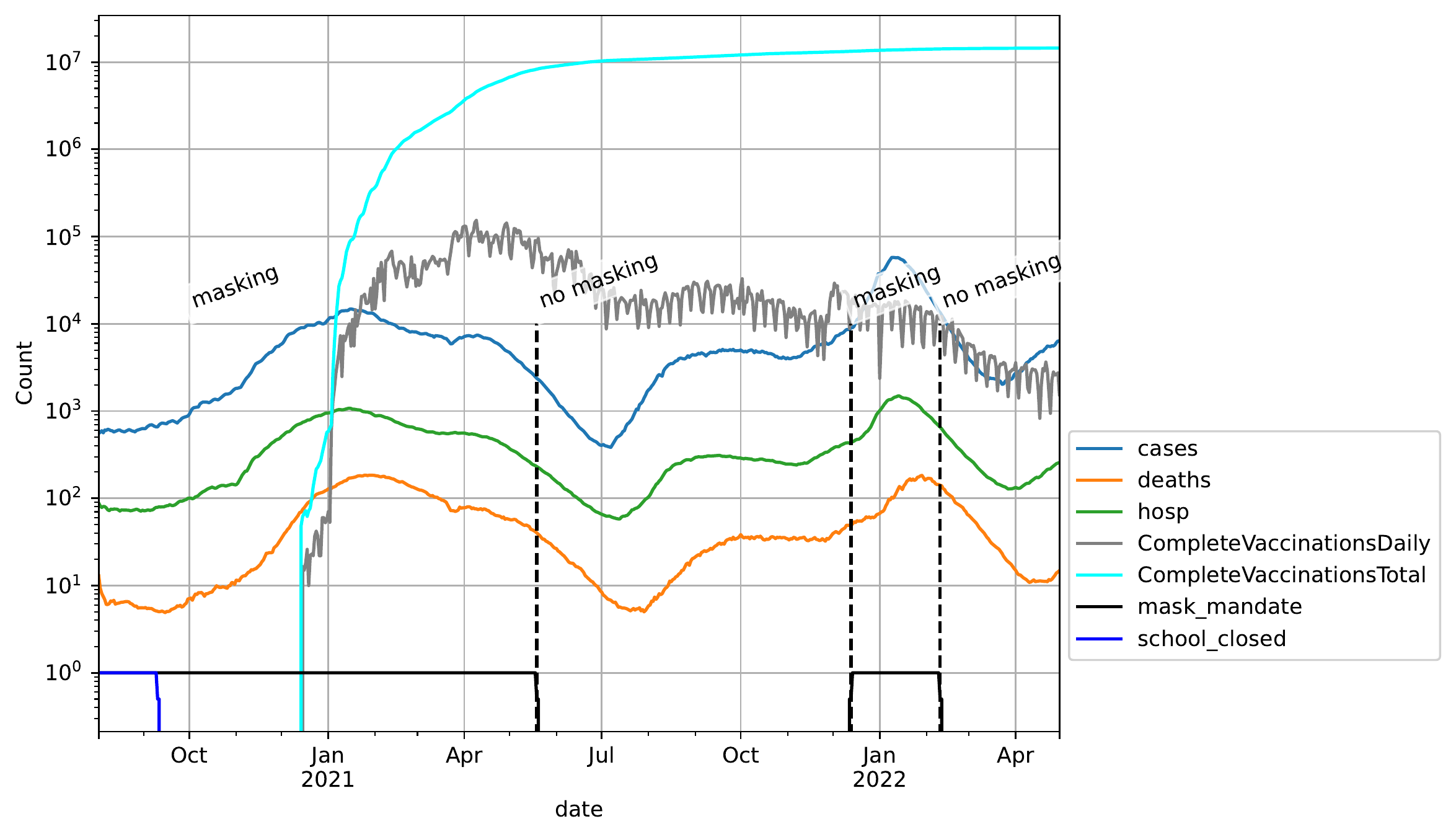}
    \caption{
        COVID-19 data used for the real world track. 
        The data consist of COVID-19 cases, deaths, and hospitalizations in New York state from August 2020 to April 30, 2022, in addition to policy indicators for masking and school closures.  
        }
    \label{fig:covid}
\end{figure}

For the real-world track, the objective was to evaluate the interpretability of SR in a real-world scenario, rather than relying on measures of complexity. 
In real-world scenarios, the interpretability of a model is subjective and depends on the expertise of specialists from the domain of application.  
To this end, we tested the ability of competitor methods to generate explainable predictive models of COVID-19 cases, hospitalizations, and deaths.

The task was constructed using time-series data for daily numbers of cases, hospitalizations, and deaths in New York state between August 2020 and April 30, 2022. 
We utilized an existing collation of these data from the Github repository \url{https://github.com/reichlab/covid19-forecast-hub/}.
In addition, we collected data on covariates hypothesized to be associated with disease outbreak, including policy measures (mask mandates, school closures) and daily and total vaccinations. 
An overview of the data trends is given in \cref{fig:covid}.

Each method was tasked with projecting future counts of cases, hospitalizations, and deaths using historical data.  

The three time-series (number of cases, hospitalizations, deaths) were pre-processed as follows. 
First, we replaced any outlier with the one week window median value. We defined as outlier any data point that was $4$ standard-deviations away from the average of one week window. After that, we generated a new time-series as the exponentially weighted moving average in order to smooth the measurements. Finally, for every day of the time-series, we extracted the following features:

\begin{itemize}
    \item $x_{t-i, j}$: the value of $i$ time steps backwards, with $j$ being the value of number of cases, hospitalization, deaths.
    \item $x_{\Delta_i, j}$: the difference between the value of $i$ time steps backwards and the value at $t=0$.
    \item $x_{tot, j}$: the cumulative sum of the values from $t=0$ until the current time step.
\end{itemize}
    
We calculated these features for $i = \{1, 2\}$ with the goal of predicting the value of $t+1$ (i.e., today's prediction is not provided).
The training and test sets were created by alternating between $5$ weeks and $3$ weeks chunks, respectively, so that the algorithms would have samples from the entire duration of the data collection.

All the algorithms were executed for $10$ times for each dataset, resulting in $10$ models per algorithm.
We selected the best model w.r.t.\ the $R^2$ score on the test set as the representative model for each candidate algorithm. 
The best models were evaluated by an infectious disease expert (Dr. Majumder) using a trust score ranging from $1$ (strong distrust) to $5$ (strong trust). 
The candidate model's final score was taken to be the harmonic mean between accuracy, simplicity, and this expert trust score.

\subsection{Competition entries}

A total of $13$ teams participated in the competition, including $1$ withdrawal. 
For brevity, all methods are summarized in Table~\ref{tab:methods}.
We report a description of each method submitted by the participants in the Supplementary Material.
We note that GP\textsubscript{ZGD} was a late submission, and was only evaluated in the qualification stage of the competition.

\begin{table*}[]
    \centering
    \caption{Summary of the methods submitted to the competition. 
    For method class, we use: EA=Evolutionary algorithm (e.g., GP), DL=Deep learning, Mix=Combination of multiple classes.}
    \begin{tabularx}{\linewidth}{lllX}
    \toprule
    \textbf{Name} & \textbf{Class} & \textbf{Code} & \textbf{Brief description}\\
    \midrule
    Bingo~\cite{randall2022bingo} & EA & \href{https://github.com/nasa/bingo}{URL} &
    \scriptsize
    Evolves general acyclic graphs with a linear representation. 
    Includes coefficient fine-tuning, algebraic simplification, and co-evolution of fitness predictors.
    \\
    E2ET~\cite{kamienny2022end} & DL &  \href{https://github.com/pakamienny/e2e_transformer}{URL} &
    \scriptsize
    A pre-trained transformer that predicts SR models directly from the data.
    Predicted models are then fine-tuned and the best is returned.
    \\
    PS-Tree~\cite{zhang2022ps} & MIX & \href{https://github.com/zhenlingcn/PS-Tree}{URL} &
    \scriptsize
    Combines decision trees, GP, and ridge regression within the evolutionary process.
    It is capable of tackling the SR problem in a piece-wise fashion.
    \\
    QLattice~\cite{brolos2021approach} & EA &\href{https://docs.abzu.ai/}{URL} & 
    \scriptsize
    Uses a probability distribution, updated over the iterations, to sample increasingly better solutions.
    It includes fine-tuning and more.
    \\
    TaylorGP~\cite{he2022taylor} & EA &\href{https://github.com/KGAE-CUP/TaylorGP}{URL} & 
    \scriptsize
    Combines GP with Taylor polynomial approximations. 
    It uses Taylor expansions to identify polynomial features and decompose the problem.
    \\
    EQL~\cite{sahoo2018learning} & DL &\href{https://al.is.mpg.de/research_projects/symbolic-regression-and-equation-learning}{URL} & 
    \scriptsize
    Consists of a fully-differentiable, shallow neural network that contains SR operators as activations.
    The $L^0$ loss is used to prune the network.
    \\
    GeneticEngine~\cite{geneticengine22} & EA &\href{https://github.com/alcides/GeneticEngine/}{URL} & 
    \scriptsize
    Uses strongly-typed and grammar-guided GP.
    For this competition, no domain-specific information was needed for the grammar.
    \\
    Operon~\cite{burlacu2020operon} & EA &\href{https://github.com/heal-research/operon}{URL} & 
    \scriptsize
    {C\raisebox{0.5ex}{\tiny\textbf{++}}}-coded GP, where fine-tuning is realized with the Levenberg-Marquardt algorithm.
    It is paired with Optuna for tuning search hyper-parameters.
    \\
    PySR~\cite{cranmer2020pysr} & EA &\href{https://github.com/MilesCranmer/PySR}{URL} & 
    \scriptsize
    Uses tree-based expressions, tournament selection, and local leaf search.
    It further uses multiple populations during the search.
    \\
    uDSR~\cite{landajuela2022a} & MIX &\href{https://github.com/brendenpetersen/deep-symbolic-optimization}{URL} & 
    \scriptsize
    Unified framework for SR that combines: recursive problem simplification, neural-guided search, large-scale pre-training, sparse linear regression, and GP.
    \\
    \texorpdfstring{GP\textsubscript{ZGD}}{gpzgd}~\citep{10.1145/3377930.3390237} & EA &\href{http://github.com/grantdick/gpzgd}{URL} & 
    \scriptsize
    Koza-style canonical GP with the addition of stochastic gradient descent to tune coefficients during the evolution.
    \\
    NSGA-DCGP & EA &\href{https://github.com/LuoYuanzhen/srbench}{URL} & 
    \scriptsize
    Combines differentiable cartesian GP~\cite{izzo2017differentiable} with the non-dominated sorting genetic algorithm II (NSGA-II) to simultaneously discover short and accurate models.
    \\
    \bottomrule
    \end{tabularx}
    \label{tab:methods}
\end{table*}

\section{Results}\label{sec:results}

This section will present a detailed analysis of the results for each track of the competition. A summary of these results can be found at the competition website\footnote{\url{https://cavalab.org/srbench/competition-2022/}}.

\subsection{Qualification Track}

\begin{figure}
    \centering
    \includegraphics[width=0.9\linewidth]{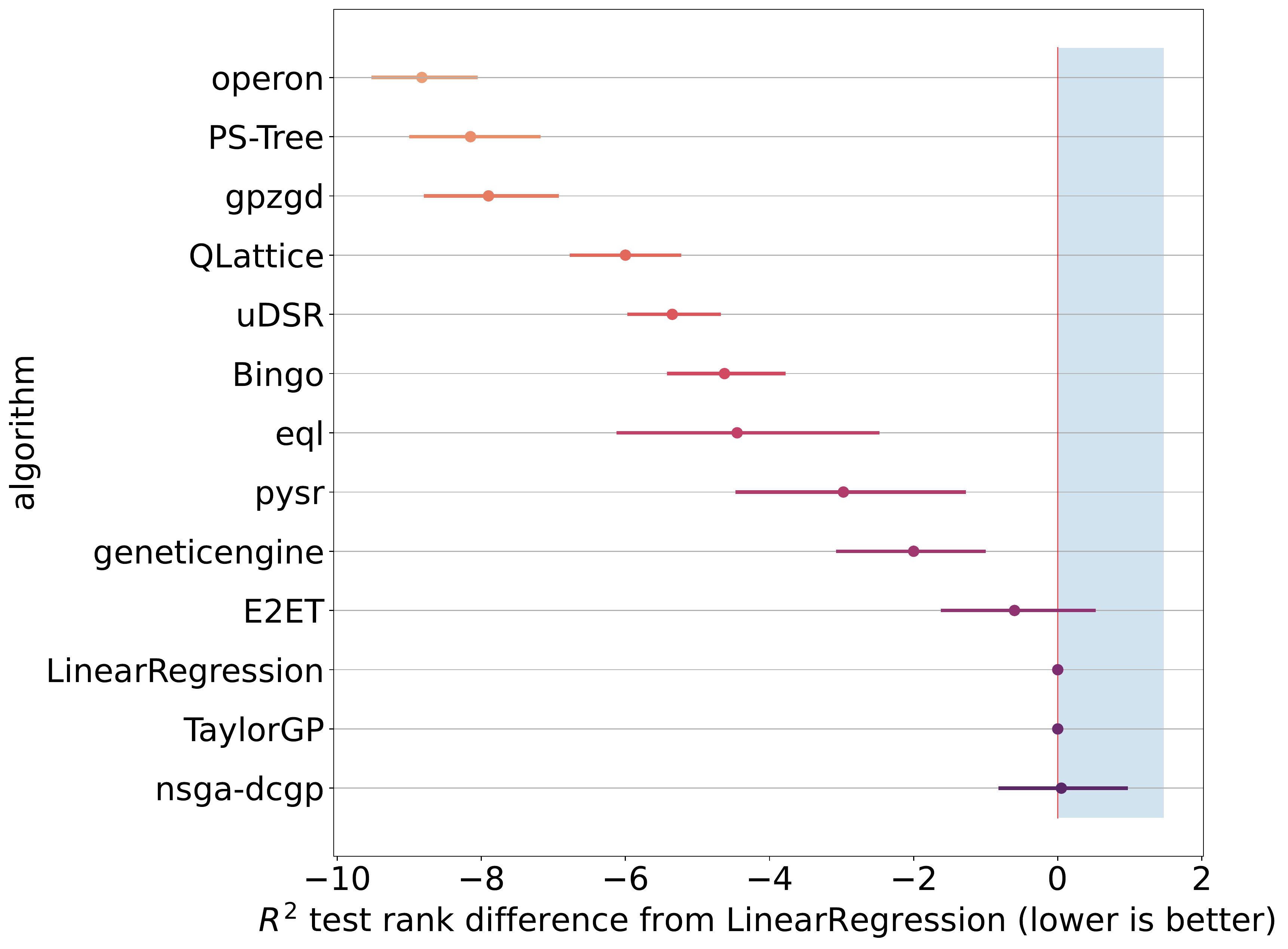}
    \caption{
        Rank of the SR methods for the qualification track calculated as the median $R^2$. 
        In this plot, the lower the value the better. 
        The shaded area depicts the disqualified competitors. 
    }
    \label{fig:quali-result}
\end{figure}

The rank of the tested SR methods for the qualification track is reported in \cref{fig:quali-result}. From this figure we can see that most of the competitors found better models than the linear regression. In this stage, the SR methods TaylorGP and nsga-dcgp were disqualified. The former simply returned linear models while the latter returned some slightly better models and some slightly worse than linear regression. Even though E2ET was better than linear at the median score, it also found a worse than linear model in some occasions. One reason being that this algorithm can only handle datasets up to $10$ variables and some of those datasets had more than that. On the other side of the spectrum, we have Operon, PS-Tree and \texorpdfstring{GP\textsubscript{ZGD}}{gpzgd} as the most accurate models for this selection of datasets.

\begin{figure}
    \centering
     \includegraphics[width=\linewidth]{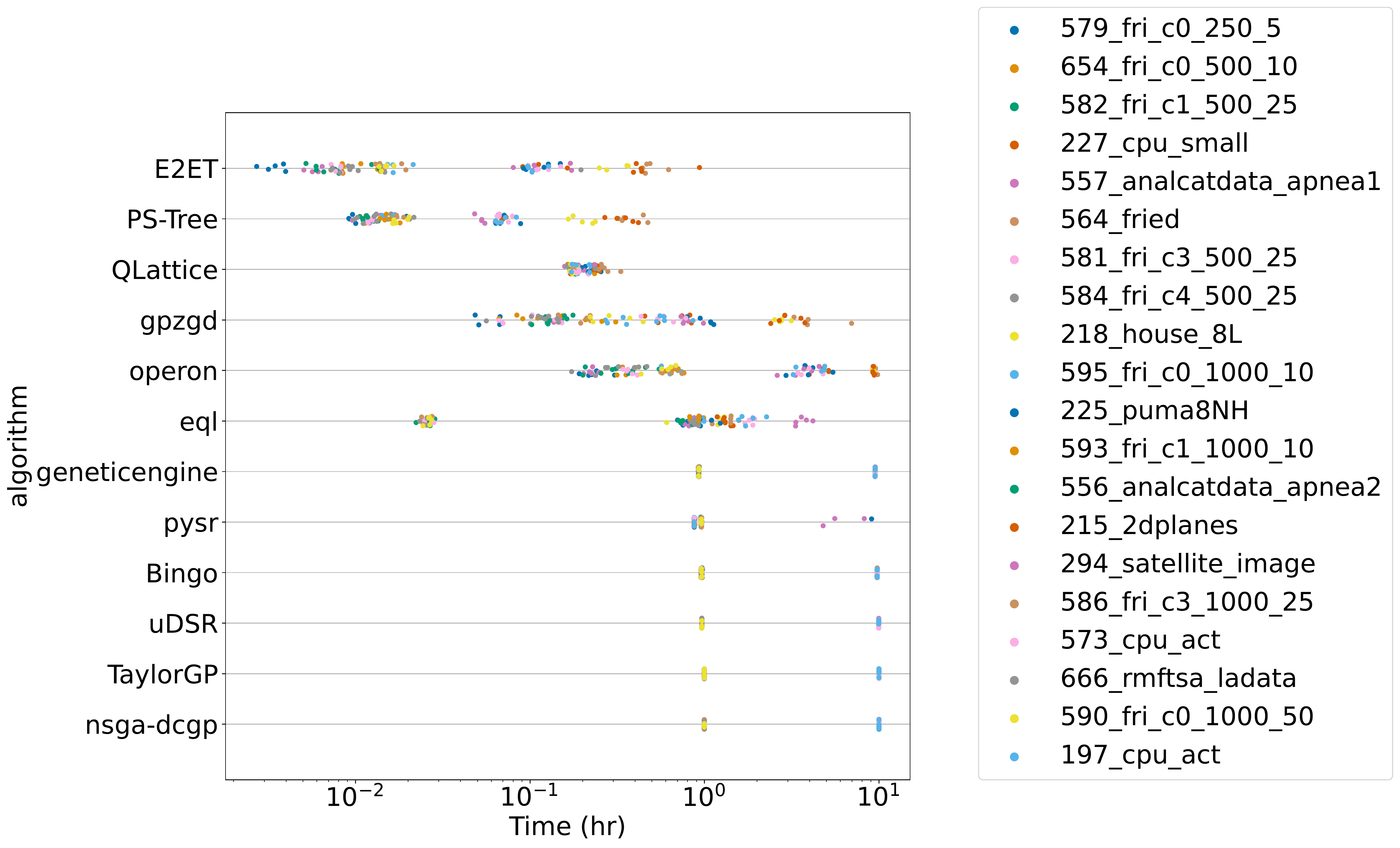}
    \caption{
        Runtime of each SR method for every tested dataset during the qualification stage.
        Note that methods were limited to one hour of total time for datasets with less than $1000$ samples and ten hours for larger datasets. 
    }
    \label{fig:runtime}
\end{figure}

Regarding the runtime, in \cref{fig:runtime} we can see that half of the participants used the full budget to find the best solution. As most of them are population-based search algorithms, it is a sane strategy to maximize the number of evaluations in the hopes of finding a good local optima. EQL, E2ET and PS-Tree adopts a different strategy to search for the expression so that there is no mechanism to stop premature convergence. As such, once it reaches a local optima, they can stop the search.

\subsection{Synthetic Benchmark Track}

\begin{figure}[t!]
    \centering
    \includegraphics[width=0.9\linewidth]{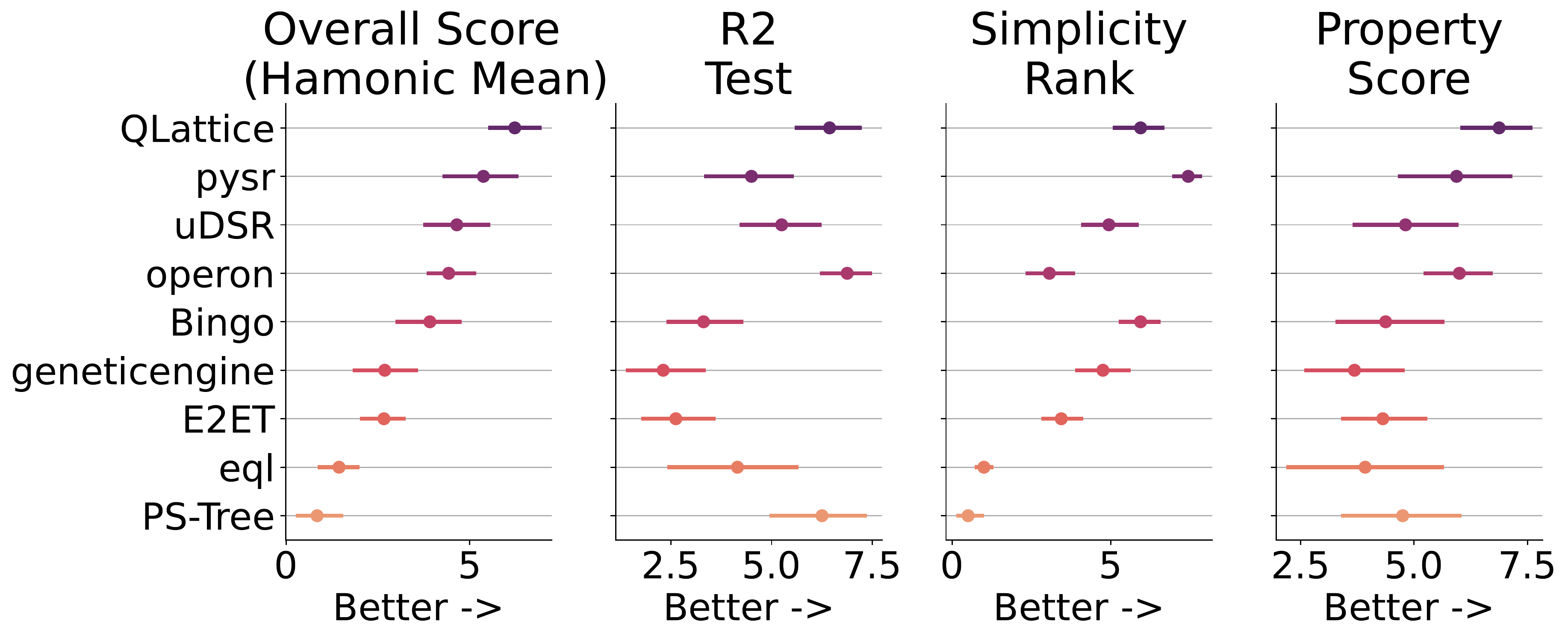}
    \caption{Aggregated results for all tasks in the Synthetic Benchmark Track. The leftmost plot presents the distribution of the harmonic mean of all three criteria, the medians of this plot was used to decide the winner. The next three plots shows the distribution of $R^2$, simplicity, and task-specific, respectivelly.}
    \label{fig:stage1-overall}
\end{figure}

Fig.~\ref{fig:stage1-overall} shows the aggregated rank comprising every task of the synthetic benchmark track. As we can see from this figure QLattice ranked first, followed by pysr and uDSR when considering the harmonic mean of all three criteria. Notice that the higher the value of the rank, the better; the reason for this is that the harmonic mean penalizes small values more, as per our intention to penalize models that does not satisfy all criteria. We can also see from this plot that there is an overlap in the distribution among the contenders, suggesting that the results may not be statistically significant.

Regarding $R^2$ score, operon returned the best models followed by QLattice and PSTree. Looking at the simplicity scores, pysr, QLattice, and Bingo were the top three, and for the property specific score the best algorithms are QLattice, pysr, and Operon. QLattice is among the top-$3$ in every criteria, corroborating with the first place in the harmonic mean of the ranks.

\begin{figure}[t!]
    \centering
    \includegraphics[width=0.7\linewidth]{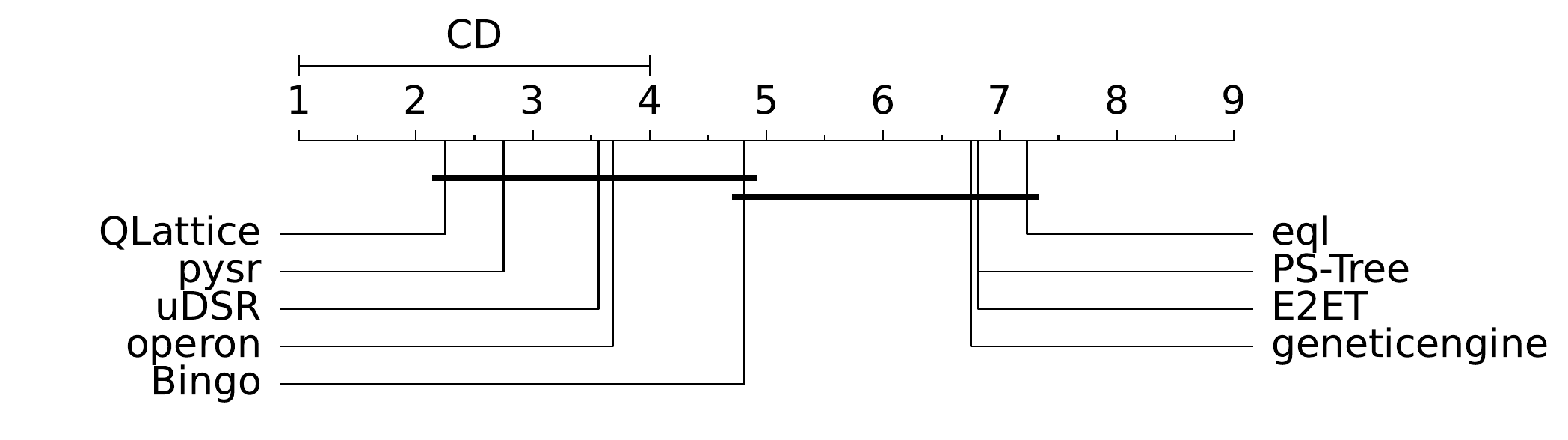}
    \caption{Critical diagram of the average of the harmonic mean of the criteria using Nemenyi test with $\alpha = 0.05$ as a post-hoc test.}
    \label{fig:stage1-overall-cd}
\end{figure}

To verify the statistical significance of the results, Fig.~\ref{fig:stage1-overall-cd}  shows the critical difference diagrams using the Nemenyi test ($\alpha = 0.05$) to find the groups of algorithms that presents a significant difference to each other. As we can observe from this diagram, the first five in rank does not have a statistical significance among their results, this means that either they are expected to perform the same on average or that we require more benchmarks to observe a significant difference.

\subsubsection{Rediscovery of exact expression}

\begin{figure}[t!]
    \centering
    \includegraphics[width=\linewidth]{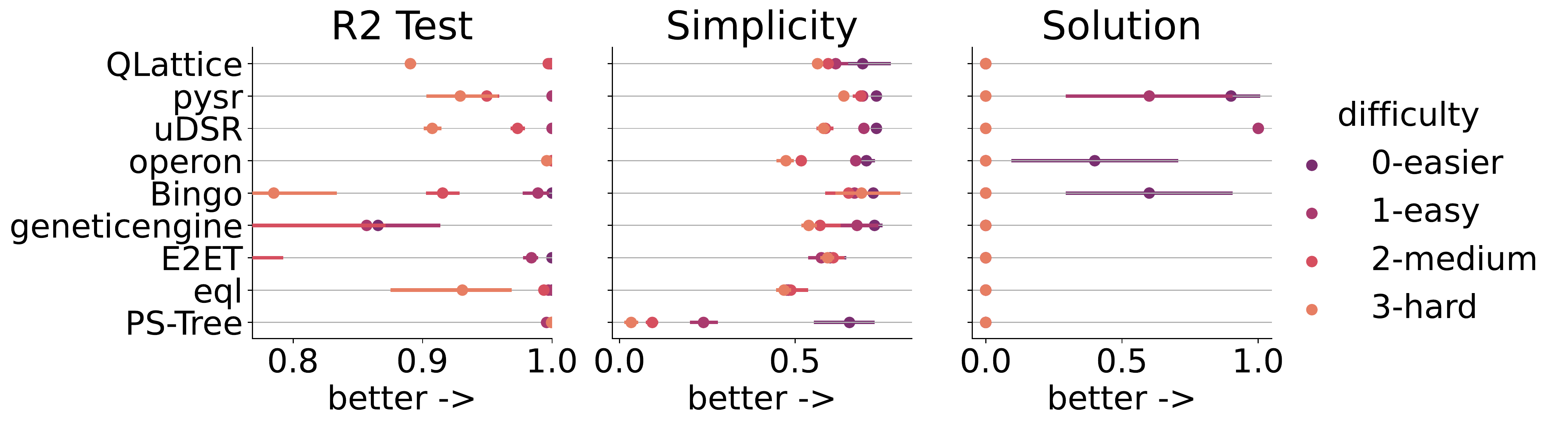}
    \caption{Distribution of ranks for the discovery of exact expressions. As per the other plots, the higher the value of the rank, the better.}
    \label{fig:stage1-exact}
\end{figure}

In Fig.~\ref{fig:stage1-exact} we can see the results for the rediscovery of exact expression task grouped by the different levels of difficulties. The first observation is that most of the algorithms are capable of finding accurate models (with $R^2$ higher than $0.9$) on every level of difficulty, the only exceptions being Bingo, geneticengine, E2ET on the higher difficulty levels. They are also capable of maintaining a similar level of simplicity in comparison to each other, except for PS-Tree that created more complex models. 

Regarding the recovery of the expression, we can see that this was only achieved for problems in the easier and easy difficulty levels. Further, only pysr, uDSR, operon, and Bingo were capable of finding exact solutions in these levels. It should be highlighted that uDSR found a perfect match in every run for those two levels. 

By selecting the model with the highest harmonic mean rank between the three criteria, we found models very close to the ground-truth obtained by Bingo and pySR. For the easier level, Bingo found the following expression:

\begin{equation*}
    f(x) = 0.4 \times (x_1 + 6.25) \times (x_2 - 3.75) + 10.37
\end{equation*}
that matches the generating function exactly after some algebraic manipulation. 


The other two difficulty levels were particularly hard for the SR algorithms. One possible reason is that SR usually have difficulties handling rational functions due to using protected division or replacing the division operator by the analytic quotient to avoid partiality. 
Using the same criteria as above, the best model returned by operon for the medium difficulty level resembled the true expression:

\begin{equation}
f(x) = \frac{- 1.66 x_{1} + \left(- 0.02 x_{1} - 0.15\right) \left(- 16.15 x_{2} - 6.46\right)}{\left(0.2 x_{1}^{2} + 1\right) \left(\frac{0.19 x_{2}^{2}}{0.2 x_{1}^{2} + 1} + 1\right)^{0.5} \left(\frac{0.21 x_{2}^{2}}{0.2 x_{1}^{2} + 1} + 1\right)^{0.5}}
\label{eq:operon-almost}
\end{equation}

If we change $0.19 x_2, 0.21 x_2$ both to $0.2 x_2$, we get an expression that, after some algebraic manipulation, becomes the true expression (\cref{tab:exact-eq}). 
 This reveals two problems: i) the internal optimization of the numerical parameters may suffer from imprecision that makes it harder to simplify the expression; ii) the evaluation criteria for the exact solution may miss correct expressions because of difficulties during the simplification process.

\subsubsection{Selection of relevant features}

\begin{figure}[t!]
    \centering
    \includegraphics[width=\linewidth]{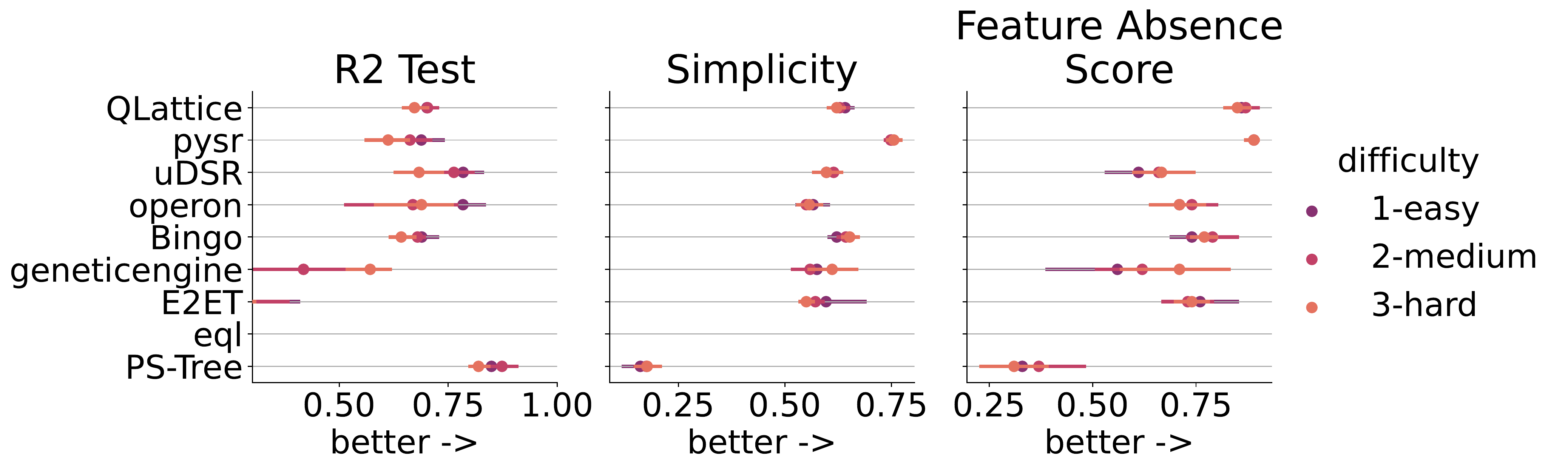}
    \caption{Distribution of ranks for the feature selection task. As per the other plots, the higher the value of the rank, the better.}
    \label{fig:stage1-featselect}
\end{figure}

Fig.~\ref{fig:stage1-featselect} shows the distribution of ranks for the task of using only the relevant features in the model. Analysing the common criteria for all tasks, we can see that PS-Tree stands out as the most accurate model and the most complex one at the same time, for every difficulty level. Incidentally, it is also the algorithm with the smallest scores for the feature absence score. Being a piecewise approach, it is likely that it created additional sub-expressions to account for the additive noise, generating a more complex expression and making use of almost every feature.


The algorithm pysr returned the model with the best harmonic mean for all three levels of difficulty. Incidentally, it was the same model for all of them:

\begin{equation*}
    f(x) = 7.1 x_{1} + x_{2} x_{3} + 0.27 x_{6}^{2} x_{7},
\end{equation*}
which is very different from the true generating function.

As we can see from this equation, the pysr model uses two irrelevant features and miss some of the true features. That being said, it is still far from a desirable solution where it captures the correct interactions between relevant features. Overall, this task has proved to be a challenge for SR models and it shows that SR can possibly benefit from the application of a feature selection approach before fitting the data.

\subsubsection{Escaping local optima}

\begin{figure}[t!]
    \centering
    \includegraphics[width=\linewidth]{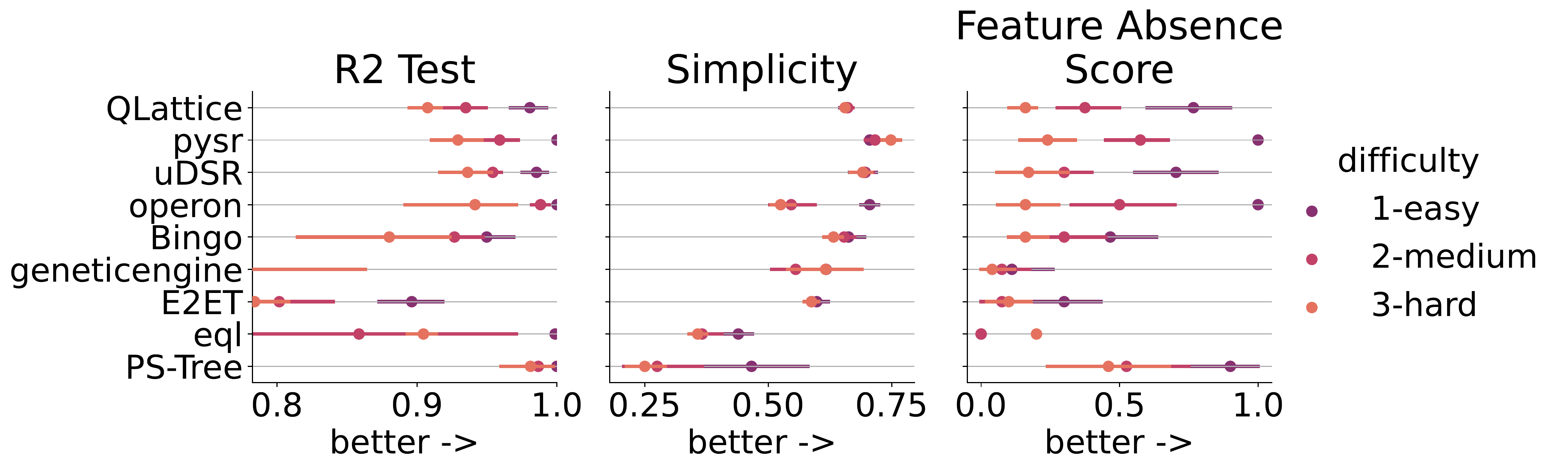}
    \caption{Distribution of ranks for the local optima task. As per the other plots, the higher the value of the rank, the better.}
    \label{fig:stage1-localopt}
\end{figure}

For the local optima task (Fig.~\ref{fig:stage1-localopt}), we can see a similar behavior of the feature selection task w.r.t.\ the accuracy and simplicity scores. For the feature absence score, we observe a more spread distribution of the results, showing that there is a significant difference between the different difficulty levels. For the easy level, pySR and Operon used only relevant features in every run. One such example being the model:

\begin{equation*}
f(x) = 3.97 x_{2} \left(0.19 x_{1} + 0.38 x_{3}\right) + 1.2 x_{4}^{2},
\end{equation*}
that corresponds to the exact generating function. For the other difficulty levels, there were no perfect solutions. The best ones following the harmonic mean are:

\begin{equation*}
    f(x) = 0.77 x_{1} x_{2} + 0.31 x_{1} x_{4} x_{5} + 1.52 x_{2} x_{3} + 1.2 x_{4}^{2},
\end{equation*}
for the medium level with operon that also corresponds to the true model and

\begin{equation*}
    f(x) = x_{10} + 1.53 x_{2} x_{3} + 1.23 x_{4}^{2} + x_{6} + x_{9},
\end{equation*}
for the hard level with pySR that uses some of the local optima features to reach an almost perfect solution. 
As we can see from these results and \cref{fig:stage1-localopt}, the noise plays an important role as it masquerades the importance of the true features. In other words, as much as pySR and Operon are competent to find the generating function, with too much noise in the target function, the noisy variables can be confused with the true variables. This task still presents a challenge and must be further investigate with more intermediate levels.

\subsubsection{Extrapolation accuracy}

\begin{figure}[t!]
    \centering
    \includegraphics[width=\linewidth]{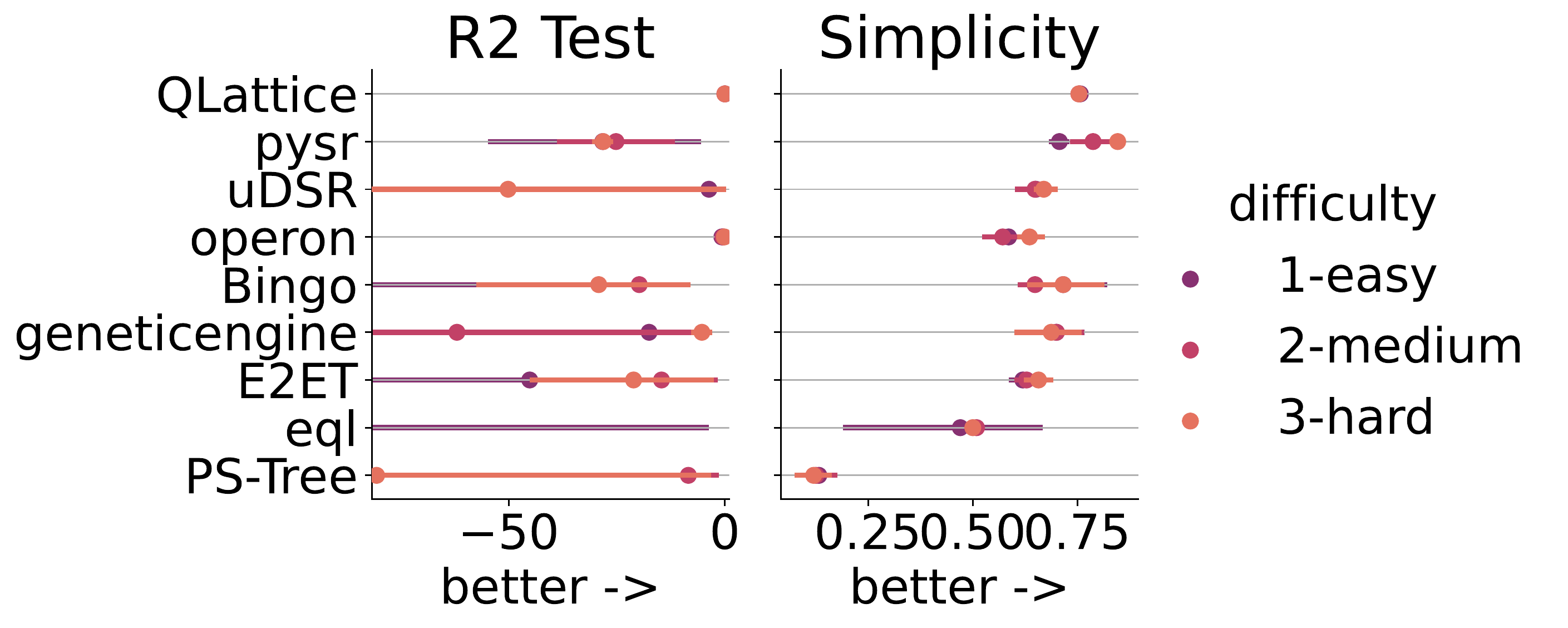}
    \caption{Distribution of ranks for the extrapolation task. As per the other plots, the higher the value of the rank, the better.}
    \label{fig:stage1-extrapolation}
\end{figure}

The extrapolation task proved to be very challenging to SR algorithms, as we can see from Fig.~\ref{fig:stage1-extrapolation}. In this plot we can see that the obtained accuracy for most of the models were subpar, often with a negative $R^2$ for the test set, even for the easiest levels. 

\begin{figure}[t!]
    \centering
    \begin{tikzpicture}\begin{axis}[domain=-15:40, ymin=-1.5, ymax=1.5, legend pos=south east,scale only axis=true,width=0.3\textwidth,height=0.3\textwidth]
      \addplot[domain=-15:40, samples=1000, olive, thick]{0.191257247217522*sin(deg(5.5016*x)) + 1.0011*tanh(0.2641*x + 0.0458257569495584*sqrt(abs(x))) - 0.0247};
      \addplot[domain=-15:40, samples=1000, red, thick]{0.333333333333333*x/sqrt(0.111111111111111*x*2 + 1)};
      \addplot[domain=-15:40, samples=1000, blue, thick]{0.989484*tanh(0.274847*x + 0.0440621) - 0.00121047};
      \addplot[domain=-15:40, samples=1000, black]{erf(0.22*x) + 0.17*sin(deg(5.5*x))};
      \addplot +[mark=none, dashed, black] coordinates {(15, -1.5) (15, 1.5)};
      \addplot +[mark=none, dashed, black] coordinates {(40, -1.5) (40, 1.5)};
      \legend{operon-easy, uDSR-medium, QLattice-hard, ground-truth}
    \end{axis}
\end{tikzpicture}
    \caption{Selected solutions for the extrapolation task with different level of difficulties. Extrapolation region is between dashed lines.}
    \label{fig:best-extrapol}
\end{figure}
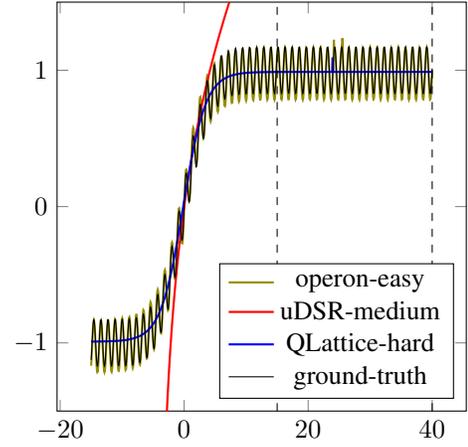

To illustrate the behavior of models from different levels, we have selected the best harmonic mean of each difficulty level as illustrate in Fig.~\ref{fig:best-extrapol}. We can see from this plot that Operon found a very close approximation to the original noiseless dataset, while QLattice returned models that approximated the error function without the sine term. The algorithm uDSR returned a function that approximated part of the interpolation region, but it was very different on the extrapolation data.

In summary, we can see that these algorithms were capable of capturing part of the target function missing one simple component. These results demand a broader investigation on why none of the algorithms were capable of including the sine function into the final expression.

\subsubsection{Sensitivity to noise}

\begin{figure}[t!]
    \centering
    \includegraphics[width=0.75\linewidth]{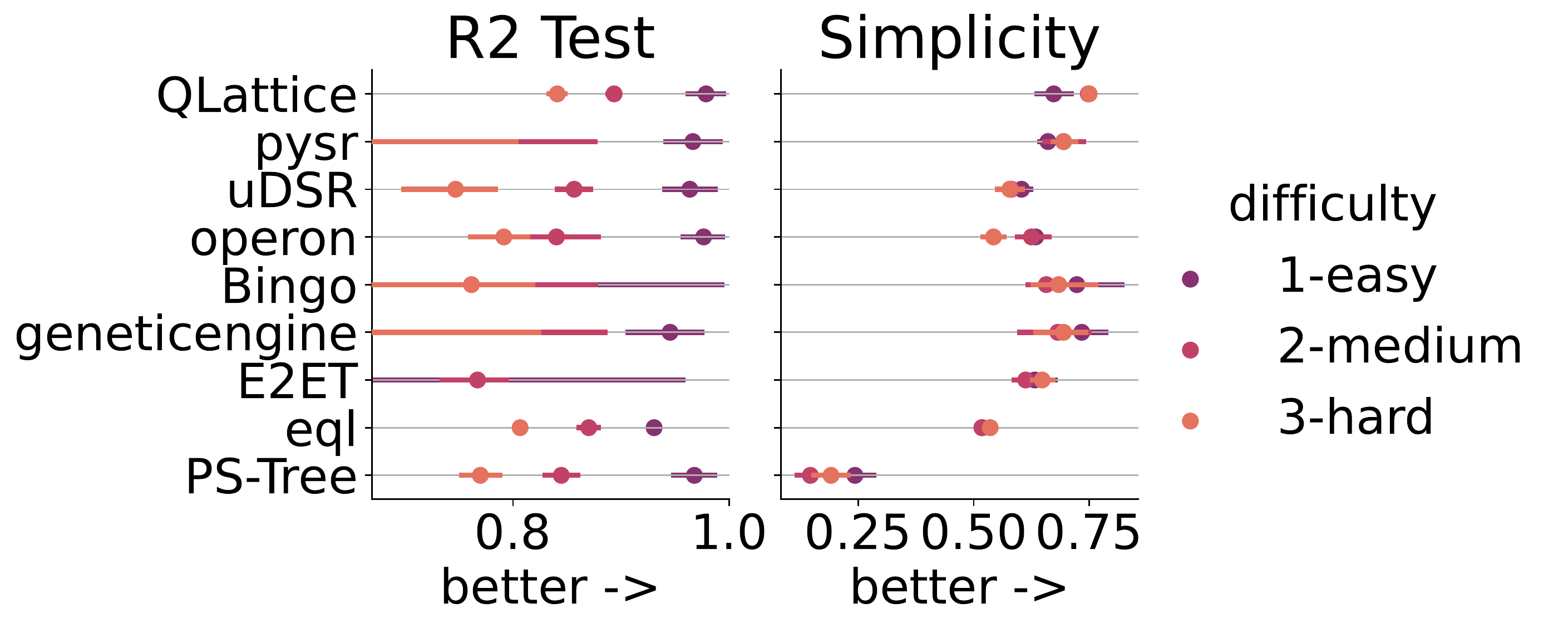}
    \caption{Distribution of ranks for the noise task. As per the other plots, the higher the value of the rank, the better.}
    \label{fig:stage1-noise}
\end{figure}

Finally, in Fig.~\ref{fig:stage1-noise} we can see the accuracy and simplicity scores for the sensitivity to noise task. From this plot we can clearly see the increase in difficulty as we increase the noise level. Even with the degradation of accuracy, almost every algorithm is capable of keeping the simplicity at a high score. Particularly for this task, QLattice, uDSR, operon, eql, and PS-Tree were capable of maintaining a relatively high accuracy across all difficulty levels.

\subsection{Real-world track}

The expert was presented with a sequence of models with the necessary information to assess their evaluation. 
The evaluation screen contained information for the task (i.e., predicting cases, hospitalization, deaths), the event horizon, algorithm name, $R^2$ score, simplicity score, the regression model expression, a plot of the true time-series with the predicted time-series, and a plot of the predicted value versus the true value. 
The expert provided us with gestalt expert rating answering the question ``I trust this model" with a scale from $1$ (strongly disagree) to $5$ (strongly agree).
The final score is the harmonic mean of the model accuracy, simplicity score, and expert score.



\begin{figure}
    \centering

    \fbox{
    \includegraphics[width=.95\columnwidth]{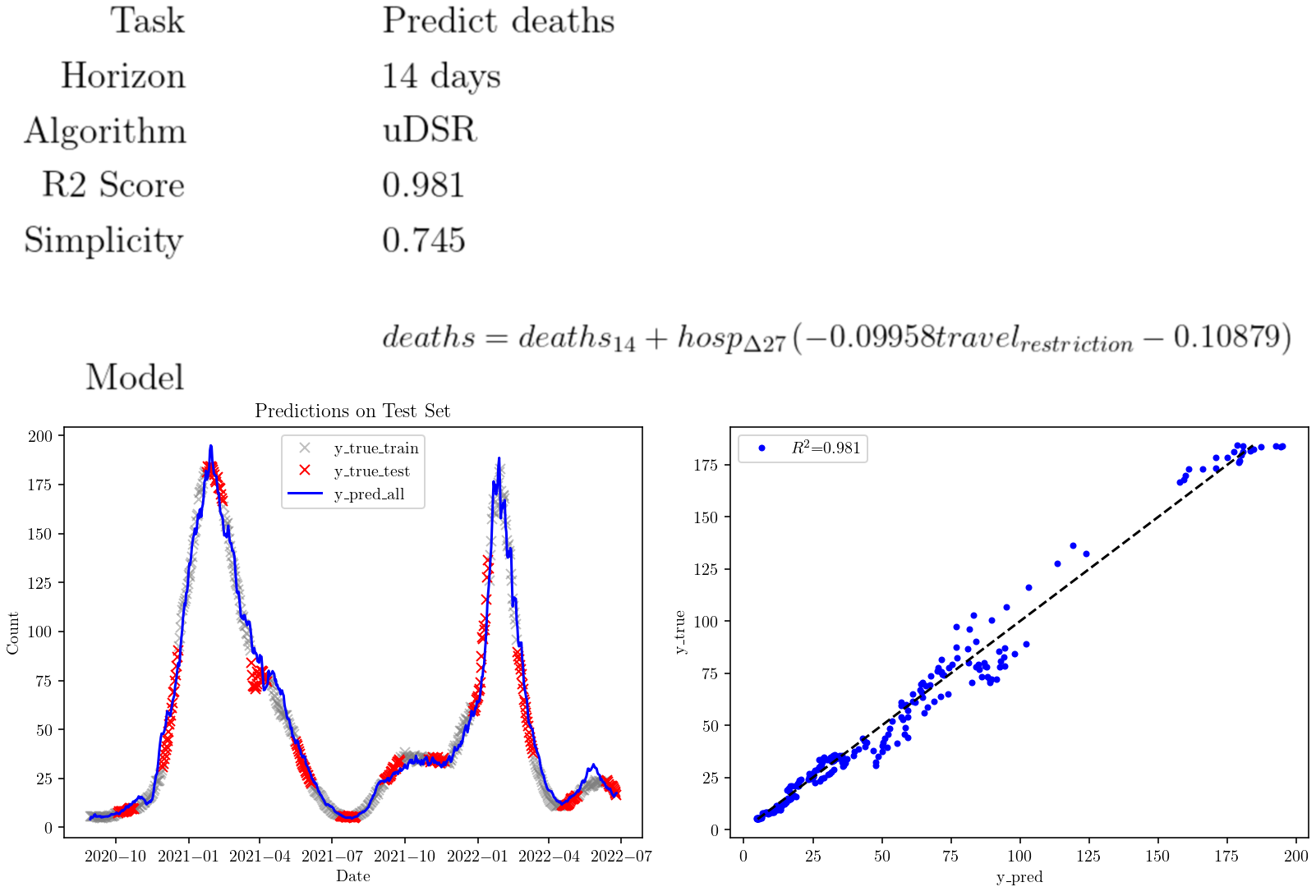}
    }
    \caption{Sample presentation of a uDSR model used for expert assessment in the real-world track. 
    }
    \label{fig:udsr}
\end{figure}

%
We report the results of this track in Table~\ref{tab:results-real-world}. 
In this track the best score was obtained by uDSR followed by QLattice and geneticengine. 
\cref{fig:udsr} shows a sample result for the prediction of the number of deaths using uDSR in the form presented to the expert for scoring. 
Although multiple approaches found good models for this task in terms of $R^2$ score, uDSR managed to return a comparably simple, plausible linear model with one interaction that generated good predictions.


\begin{table}[t!]
    \centering
    \caption{Results for the real-world track.}
    \begin{tabular}{@{} >{$}c<{$}|c|>{$}c<{$}|>{$}c<{$}|c|>{$}c<{$} @{}}
    \textbf{Rank} & \textbf{Algorithm} & \textbf{Score} & \textbf{Rank} & \textbf{Algorithm} & \textbf{Score} \\
    \hline 
    1  & uDSR & $5.75$ & 5  & Bingo & $4.66$\\
    2  & QLattice & $5.21$ & 6  & pySR & $4.17$ \\
    3  & geneticengine & $4.99$ & 7  & PS-Tree & $3.15$\\
    4  & Operon & $4.80$ & 8  & E2ET & $2.72$\\
    \hline
    \end{tabular}
    \label{tab:results-real-world}
\end{table}

\section{Discussion}\label{sec:discussion}

Overall, this competition provided many insights about the current state of SR. 
In the remainder of this section, we focus on discussing different aspects of the competition itself.
Throughout these benchmarks we noticed that there are still some gaps on the different degrees of difficulty for each task, filling up these gaps is important to highlight the differences between algorithms and their limitations.

\subsection{Limitations of the competition}

Regarding the limitations of this competition, one observation is that some algorithms performed better on one track and worse on another, or even they were better when observing one of the criteria and worse on another. Even though we tried to balance this by using the harmonic mean, an aggregated final score hides the fine details of the results.
Looking at the synthetic tracks, we can see that some algorithms are tweaked toward a better accuracy, like Operon and QLattice, while others tried to balance the accuracy-simplicity tradeoff, like pysr. As some of these approaches employ different mechanisms to find a balance, would a better tweaking of the hyperparameters render a better result for those algorithms?


The simplicity measure only takes into account the size of the expression, but there are some constructs that hinders the interpretability of an expression without adding too much to the size. Consider the chain of nonlinear functions, like $f(g(h(x))$ that accounts for $4$ nodes and a simpler expression like $f(x) + h(x)*g(x)$, that accounts for $8$ nodes. Even though the first one is smaller, the second one can be easier to reason about its behavior than the first. There are some examples of complexity measures such as in~\cite{kommenda2015complexity,virgolin2020learning}.

Another issue is the computational budget, since we have a limited resource to run the experiments we have to limit the the maximum runtime and repetitions. While the established runtime can be considered reasonable, a large number of repetitions could reduce some uncertainties about the results. 

Finally, we should stress that there is hardly a single winner or a single approach that could be named state-of-the-art. We have to use an aggregation score to elect the winner; however, inspecting the results we can see that it is often the case that there are different winners on each track and each level.

\subsection{Overall difficulty of the tasks}

Assessing the difficulty of the tasks can be challenging since it is not possible to test the problem instances before the competition due to the risk of biasing the results. As it turns out, some of the tasks proved to be really challenging. 

For example, in the task of exact solution, just a few correct expressions were found and only on the easier and easy difficulties. This has two main reasons: i) any expression can be written using different alternative expressions such that the simplification procedure fails to match them; ii) the optimization of numerical parameters can lead to imprecision that causes the expression to be close but not equal to the ground-truth (as in Eq.~\ref{eq:operon-almost}).

The extrapolation task was also difficult for all of the approaches that approximated the error function well but without the sine term of the ground truth. On one hand, given that a certain dataset can have many different solutions inside the interpolation region that behaves differently outside this region, we cannot expect the algorithm to make the right choice without any prior information. On the other hand, none of the algorithms could find a good solution even for the training data, revealing that the generating function was already challenging. 

Finally, the local optima and sensitivity to noise tasks provided a good balance between difficulties allowing us to verify that, while current SR algorithms can find a good solution with low noise data, the effect of the noise can have a large impact on the quality of the model. In the local optima task, we can see that the algorithms made a preference of maximizing simplicity even if that adds noise to the expression, this can be expected as it is an informed choice of choosing the expression with the smaller number of nodes even if that reduces the accuracy a bit.

Overall, it is important that we continuously adapt the benchmark problems using the information gained by past competitions to improve the difficulty range without introducing any bias towards any of the competitors. Additionally, a continuous adaptation would serve to avoid that the SR approaches evolve towards beating these specific benchmarks, a common problem in many fields.

\subsection{Manual inspection is still a challenge}

About the real-world track, there are still some challenges that should be assessed. First of all, the expert in the field had to evaluate many different models by looking at the analytic expression (whenever it had a reasonable size), the interpolation and extrapolation behavior of the time series, and the accuracy score. Comparing two or more models can already be challenging even for an expert in the field, as it requires them to first establish what is an interpretable model for them. Sometimes this ad-hoc evaluation may leads to some distortions on the result since there is no way to make a pairwise comparison between models without a quadratic increase the complexity of the evaluation task.


\subsection{Are we there yet?}

In close, with a broad perspective of the results we can say, while we have witnessed many advances in this field, there are still some open challenges for the SR community.

One challenge is the support for a customised experience to the end-user, for example, specifying the degree of importance for a simpler solution when you have less accurate alternatives. In this same end, having alternative models that have a similar training accuracy could help the user to decide which one has the most probable extrapolation behavior. Additionally, as much as an analytic solution can be considered more interpretable than an opaque model, the SR tool should provide supporting information in the form of uncertainties quantification, visual behavior, and additional data such as partial derivatives and properties of the model. A minor issue is the runtime of these algorithms, as they mostly rely on evolutionary algorithms, they are usually slower than common regression algorithms.

Specifically on the assessed tasks, there are still room for improvement on the treatment of noise and feature selection. While most algorithms were capable of adequately handling noise and to remove some of the not useful features, the noise data still had an impact on the accuracy reducing the $R^2$ by up to $20\%$. Also, the removal of not needed features can lead to even simpler models that are more informative to the end-user.


\section{Conclusion}\label{sec:conclusion}

This paper provided an in-depth analysis of the 2022 Interpretable Symbolic Regression for Data
Science competition held at GECCO 2022. The main purpose of the competition was to understand how modern SR algorithms handle common challenges in data science for regression analysis and to have an indication of where the current state-of-the-art stands.

For this purpose we have created different tasks divided in separate tracks: i) rediscovery of exact expressions; ii) feature selection; iii) avoiding noisy local optima; iv) accuracy in extrapolation; v) noisy data. All these tasks are relevant not only to regression analysis but also for the interpretability of the models. We also held an additional track with the evaluation of the generated models on real-world data by an expert in their respective field.

While the competition has elected one winner for the synthetic data and another for the real-world track, we can see that there is no dominating algorithm that returns the best model in every criteria. These results help to understand the advantage and disadvantages of each approach and it can move the field forward to better algorithms.

In the near future we will release new editions of this competition with new and improved tasks and benchmark following the experience acquired during this first edition. We will also use different evaluation criteria that can help us to better understand how each algorithm stands in regarding to the different challenges.

\section*{Acknowledgment}

The authors would like to thank all of the competitors for providing helpful feedback about the competition. 
The competition was sponsored by the Computational Health Informatics Program at Boston Children's Hospital and the Heuristic and Evolutionary Algorithms Laboratory (HEAL) at the University of Applied Sciences upper Austria.
F.O. de Franca was supported by Funda\c{c}\~{a}o de Amparo \`{a} Pesquisa do Estado de S\~{a}o Paulo (FAPESP), grant number 2021/12706-1.
G. Espada, L. Inglese, and A. Fonseca were supported by \textit{Fundação para a Ciência e Tecnologia} (FCT) UIDB/00408/2020, UIDP/00408/2020, SFRH/BD/137062/2018, UI/BD/151179/2021 and CMU--Portugal Dual Degree PhD Program (SFRH/BD/151469/2021), CMU--Portugal project CAMELOT (LISBOA-01-0247-FEDER-045915), RAP project under the reference (EXPL/CCI-COM/1306/2021).
M. Landajuela, B. Petersen, R. Glatt, N. Mundhenk, and C.S. Lee were supported by Laboratory Directed Research and Development project 19-DR-003. Their work was performed under the auspices of the U.S. Department of Energy by Lawrence Livermore National Laboratory under contract DE-AC52-07NA27344. Lawrence Livermore National Security, LLC. LLNL-JRNL-846810. 
W.G. {La Cava} was supported by National Institutes of Health grant R00-LM012926.
M.S. Majumder was supported in part by grant R35GM146974 from the National Institute of General Medical Sciences, National Institutes of Health. The funder had no role in study design, data collection and analysis, decision to publish, or preparation of the manuscript.

\section*{Declarations}

F. O. de Franca, M. Virgolin, M. Kommenda, and W. G. La Cava designed and conducted the competition and analyzed the results. 
M. S. Majumder served as domain expert for the real-world track. 
The remaining authors participated in the competition and contributed to this manuscript after results were announced. 
The authors have no conflicts of interest to declare that are relevant to the content of this article.

\bibliographystyle{IEEEtran}
\bibliography{bibliography} 

\appendix

\section{Participation instructions}~\label{sec:participation}

In order to participate, the candidates had to provide an installation script that could be run without administrative permissions on a Linux system inside a \emph{conda}~\footnote{\url{https://www.anaconda.com/}} environment~\cite{anaconda}, a \emph{metadata.yml} document describing their submission, and a Python script called \emph{regression.py} containing a \emph{scikit-learn}~\footnote{\url{https://scikit-learn.org/}} Regressor object~\cite{pedregosa2011scikit}, and a function returning a \emph{sympy}\footnote{\url{https://www.sympy.org/en/index.html}}~\cite{sympy} compatible string of the SR model.
For each run, a candidate SR algorithm had a pre-specified time budget of $1$ hour for datasets up to $1000$ samples and $10$ hours for datasets up to $10000$ samples. Candidates were  responsible for ensuring that the runtime of their algorithm (including choices of hyperparameter optimization) would not exceed the budget.
In order to make comparisons as fair as possible, participants were not given advance notice of what datasets would be used for the competition. 
They knew only that datasets would follow the format in the Penn Machine Learning Benchmark (PMLB)~\cite{romanoPMLBV1Opensource,olsonPMLBLargeBenchmark2017}, and that datasets from PMLB would be used in the qualifying stage. 

With the objective of making the competition accessible and reproducible by external peers, we created a new branch in the SRBench repository\footnote{\url{https://github.com/cavalab/srbench/tree/Competition2022}} with instructions on how to add a new method and how to run the competition on a local machine.
Following these requirements, it is possible for the general public to reproduce the competition and experiment with these SR algorithms by cloning the competition branch and following the installation and execution instructions.

\section{Datasets of the qualification stage}

Table~\ref{tab:qualifying-datasets} shows the selected datasets used in the qualification stage. They were chosen as a subset of the PMLB datasets that the top-$10$ algorithms from srbench performed equally well.

\begin{table}[t!]
    \centering
    \caption{Datasets used during the qualification stage.}
    \begin{tabular}{@{} c|>{$}c<{$}|>{$}c<{$} @{}}
     \textbf{dataset} & \textbf{\# samples} & \textbf{\# variables} \\
     \hline
     197\_cpu\_act & 8192 &	22 \\
     215\_2dplanes & 40768 & 11 \\
     227\_cpu\_small & 8192 & 13 \\
     556\_analcatdata\_apnea2 & 475 & 4 \\
     557\_analcatdata\_apnea1 & 475 & 4 \\
     564\_fried & 40768 & 11 \\
     573\_cpu\_act & 8192 & 22 \\
     218\_house\_8L & 22784 & 9 \\
     225\_puma8NH & 8192 & 9 \\
     294\_satellite\_image & 6435 & 37 \\
     666\_rmftsa\_ladata & 508 & 11 \\
     579\_fri\_c0\_250\_5 & 250 & 6 \\
     586\_fri\_c3\_1000\_25 & 1000 & 26\\
     590\_fri\_c0\_1000\_50 & 1000 & 51\\
     593\_fri\_c1\_1000\_10 & 1000 & 11 \\
     595\_fri\_c0\_1000\_10 & 1000 & 11 \\
     654\_fri\_c0\_500\_10 & 500 & 11 \\
     581\_fri\_c3\_500\_25 & 500 & 26 \\
     582\_fri\_c1\_500\_25 & 500 & 26 \\
     584\_fri\_c4\_500\_25 & 500 & 26 \\
     \hline
    \end{tabular}
    \label{tab:qualifying-datasets}
\end{table}

\section{Ranking}

As we used multiple evaluation criteria for each track: $R^2$, simplicity, and a task-specific score for each tasks, we computed the aggregated rank as the harmonic means of the ranks for each criterion for $n$ different data-sets.

\begin{equation}
    \operatorname{agg\_rank} = \frac{n}{\sum_{i=1}^{n}{\frac{1}{\operatorname{rank}_i}}}
\end{equation} 
The harmonic mean imposes that, to be highly ranked, you cannot have a low rank in any of the criteria. 
This avoids the situation that an SR algorithm returns a very simple model with low $R^2$ and still ranks high among the competitors.

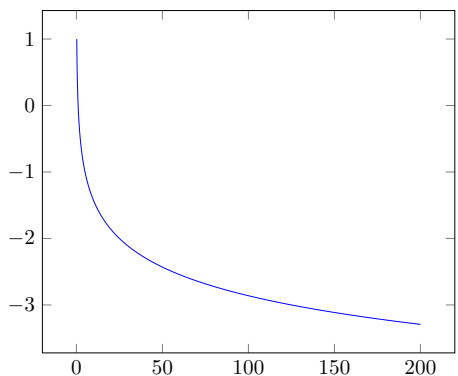
\begin{figure}[h!]
    \centering
    \begin{tikzpicture}[scale=0.8]\begin{axis}[domain=0:200]
      \addplot[domain=0:200, samples=1000,blue] gnuplot[id=erf]{-log(x)/log(5)};
    \end{axis}
\end{tikzpicture}
    \caption{Plot of the simplicity score.}
    \label{fig:simpl_score}
\end{figure}

The simplicity score is defined as $\operatorname{round}(-\log_5(s), 1)$, where $s$ is the number of nodes in the expression tree after being simplified by \emph{sympy}~\cite{sympy}, and $\operatorname{round}(x, n)$ rounds the value $x$ to the $n$-th place.
Rounding was introduced to provide some tolerance for similarly-sized expressions. 
\cref{fig:simpl_score} demonstrates how this simplicity score relates to the number of nodes in a tree. 

\section{Description of the methods}
\label{apdx:methods}

\textbf{Bingo~\cite{randall2022bingo}} implements a GP strategy for SR with a few distinctions from typical generic-programming-based SR.  First, expressions are represented as acyclic graphs, rather than trees.  Fitness calculations of the expressions involves a two step procedure: the acyclic graphs are algebraically simplified , and a gradient-based local optimization is performed on numeric constants. Lastly, coevolution of fitness predictors \cite{schmidt2008coevolution} is used for large datasets ($>1200$ datapoints). The several hyperparameters of the method were tuned prior to submission (details in \cite{randall2022bingo}) to place emphasis on obtaining  low-complexity models.  
    
     \textbf{E2ET~\cite{kamienny2022end}} involves utilizing a generative model to generate analytical expressions based on given observations. Specifically, a transformer-based neural network that is conditioned on the observations, generates a distribution over expressions. The process of sampling from this distribution involves utilizing auto-regressive decoding of expression symbols, which includes variables, operator tokens, and tokenized floats. To further enhance the generated expressions, the constants are optimized using the BFGS algorithm. This search process is highly efficient and takes only approximately one second for a single forward-pass.
    
     \textbf{PS-Tree~\cite{zhang2022ps}}, or Piecewise SR Tree, is a piecewise regression method that combines decision trees, GP, and ridge regression to divide the feature space to multiple subspaces and create a set of nonlinear models in subspaces for making prediction. The proposed PS-Tree aims to achieve high-predictive performance while retaining a reasonable level of interpretability. Specifically, PS-Tree evolves a set of GP trees under a multi-objective optimization framework to evolve a common set of non-lienar features for all ridge regression models in each subspaces simultaneously. Then, it dynamically constructs a decision tree during the evolution process to automatically determine the optimal partition scheme based on the runtime information. By doing so, PS-Tree is able to achieve high accuracy for solving piecewise regression tasks with reasonable interpretability.
    
     \textbf{QLattice~\cite{brolos2021approach}} is a framework for SR that mixes GP and numerical optimization. It is accessible through the Python library feyn. The QLattice manages a pool of models, and samples new ones via edits on the best performing models in the pool. Model performance is only evaluated after optimizing the free parameters in each model by an internally-tuned gradient descent algorithm on the training data. The available edit operations on models are chosen among by means of weighted sampling, where the weights have been optimized by Abzu. Training data with categorical features is also inherently supported, where each possible category is assigned a weight arrived at through fitting the training data. In this competition, the best model as selected by the Bayesian Information Criterion was returned from the final set of models.
    
     \textbf{TaylorGP~\cite{he2022taylor}} combines SR with Taylor polynomial approximations.
    The method uses Taylor expansions both to identify polynomial features and to decompose problems into simpler tasks.
    
     \textbf{EQL~\cite{sahoo2018learning}} is a fully differentiable neural network for SR, which permits it to be used in contexts requiring differentiable end-to-end computations. It works by approximating the equation by a shallow network with non-linear activation functions, which can be adapted to the task at hand. While sparsity can be achieved through different means, for this competition we optimized the stochastic $L^0$ loss of the expansion parameters and selected the best expression based on the quality of the fit and the simplicity of the expression according to Pareto optimality. Once the expression is fixed, the appearing constants are finally optimized via BFGS. 

     \textbf{GeneticEngine~\cite{geneticengine22}}  is a Python framework that supports different flavours of Strongly-Typed and Grammar-Guided GP. The entry used a tree-based representation, as it produces higher quality crossovers than array based approaches . Grammars are defined using standard Python classes and inheritance, allowing domain-specific knowledge to be easily incorporated into the search process. While this is the main advantage of the approach, in this competition no domain-specific information was used in the grammar. 
     
     \textbf{Operon~\cite{burlacu2020operon}} is a {C\raisebox{0.5ex}{\tiny\textbf{++}}} framework for SR with a focus on performance and scalability. It supports single- and multi-objective tree-based GP and local tuning of model coefficients via the Levenberg-Marquardt algorithm. Parallelism is implemented at the level of individual recombination events using a flexible graph-based task model~\cite{huangTaskflow2022}. Additionally, data-level parallelism is employed during tree evaluation. The Python module, PyOperon, makes this functionality available as a scikit-learn estimator. For this competition, Operon is paired with Optuna~\cite{optuna_2019} for hyperparameter tuning the best model from the last-generation Pareto front is returned. 
     
     \textbf{PySR~\cite{cranmer2020pysr}} and its accompanying backend, SymbolicRegression.jl, is an evolutionary optimization algorithm for tree-based expressions based on tournament selection and local leaf searches. PySR emphasizes performance and is very configurable---it does not place a particular prior over expression space, or define any standard operators or functional forms. PySR achieves efficient evaluation by JIT-compiling operators into fused SIMD kernels, and by multi-threading at the population-level. The full Pareto front of expressions is recorded during the search, but for this competition, the equation with the greatest improvement in accuracy along the Pareto front is returned.
    
     \textbf{uDSR~\cite{landajuela2022a}}, or Unified Deep SR, is a modular, unified framework for SR that combines five solution strategies: recursive problem simplification \citep{udrescu2020ai}, neural-guided search \citep{petersen2019deep}, large-scale pre-training for problem generalization \citep{biggio2021neural}, sparse linear regression using nonlinear basis functions \citep{brunton2016discovering}, and combination of evolutionary search with learning \citep{mundhenk2021symbolic}. By viewing these modules as connected but non-overlapping components within an algorithmic framework, uDSR is able to harness their unique strengths and address their respective limitations. The modular setup even works for a subset of the proposed modules and allows for the easy exchange of individual modules with new methods should they become available. uDSR is implemented using the open-source Deep Symbolic Optimization framework \footnote{\url{https://github.com/brendenpetersen/deep-symbolic-optimization}}.
     
     \textbf{\texorpdfstring{GP\textsubscript{ZGD}}{gpzgd}~\citep{10.1145/3377930.3390237}} is a straightforward `out-of-the-box' implementation of Koza-style canonical GP~\citep{koza1992genetic} for SR, with the addition of stochastic gradient descent (SGD) applied to model coefficients during evolution. To make the SGD more effective, Z-score standardisation is applied to all regressors. Hyperparameter tuning over Koza's canonical settings was performed to optimise performance to SR as outlined in~\citep{10.1145/3520304.3534040}. \texorpdfstring{GP\textsubscript{ZGD}}{gpzgd} was a late submission to the competition, so it was only assessed in the qualification stage of the competition.
     
     \textbf{NSGA-DCGP~\cite{izzo2017differentiable}} is a method combining differentiable cartesian GP~\cite{izzo2017differentiable} with NSGA-II with the objectives of mimizing error and complexity. 

\section{Data Availability} 

The full results files of the competition are available from the following url: \url{https://zenodo.org/record/6842176}. 
The code to run the analysis is available from \url{https://cavalab.org/srbench/competition-guide/}.



\end{document}